\documentclass[10pt,conference]{IEEEtran}

\usepackage{cite}

\ifCLASSINFOpdf
   \usepackage[pdftex]{graphicx}
   \graphicspath{{figs/}}
   \DeclareGraphicsExtensions{.pdf,.jpeg,.png}
\else
   \usepackage[dvips]{graphicx}
   \graphicspath{{../figs/}}
   \DeclareGraphicsExtensions{.eps}
\fi
\usepackage[cmex10]{amsmath}
\usepackage{amsthm}

\usepackage[ruled,vlined]{algorithm2e}

\usepackage{array}

\ifCLASSOPTIONcompsoc
  \usepackage[caption=false,font=normalsize,labelfont=sf,textfont=sf]{subfig}
\else
  \usepackage[caption=false,font=footnotesize]{subfig}
\fi
\usepackage{balance, multirow, url}

\usepackage{booktabs}
\newcommand{\ie}{\textit{i.e.}}
\newcommand{\eg}{\textit{e.g.}}
\newcommand{\etal}{\textit{et~al.}}

\begin{document}
%
\thispagestyle{empty}
\onecolumn
\linespread{1.2}\selectfont{}
{\noindent\Huge IEEE Copyright Notice}\\[1pt]

{\noindent\large Copyright (c) 2021 IEEE

\noindent Personal use of this material is permitted. Permission from IEEE must be obtained for all other uses, in any current or future media, including reprinting/republishing this material for advertising or promotional purposes, creating new collective works, for resale or redistribution to servers or lists, or reuse of any copyrighted component of this work in other works.}\\[1em]

{\noindent\Large Accepted to be published in: 2021 34th SIBGRAPI Conference on Graphics, Patterns and Images (SIBGRAPI'21), October 18--22, 2021.}\\[1in]

{\noindent\large Cite as:}\\[1pt]

{\setlength{\fboxrule}{1pt}
 \fbox{\parbox{0.65\textwidth}{L. F. A. Silva, D. C. G. Pedronette, F. A. Faria, J. P. Papa, and J. Almeida, ``Improving Transferability of Domain Adaptation Networks Through Domain Alignment Layers'' in \emph{2021 34th SIBGRAPI Conference on Graphics, Patterns and Images (SIBGRAPI)}, Gramado, RS, Brazil, 2021, pp. 168--175, doi: 10.1109/SIBGRAPI54419.2021.00031}}}\\[1in]
 
{\noindent\large BibTeX:}\\[1pt]

{\setlength{\fboxrule}{1pt}
 \fbox{\parbox{0.95\textwidth}{
 @InProceedings\{SIBGRAPI\_2021\_Silva,
 
 \begin{tabular}{lll}
  & author    & = \{L. F. A. \{Silva\} and 
                    D. C. G. \{Pedronette\} and
                    F. A. \{Faria\} and
                    J. P. \{Papa\} and
                    J. \{Almeida\}\},\\
			   
  & title     & = \{Improving Transferability of Domain Adaptation Networks Through Domain Alignment Layers\},\\
			   
  & pages     & = \{168--175\},\\
  
  & booktitle & = \{2021 34th {SIBGRAPI} Conference on Graphics, Patterns and Images ({SIBGRAPI})\},\\
  
  & address   & = \{Gramado, RS, Brazil\},\\
  
  & month     & = \{October 18--22\},\\
  
  & year      & = \{2021\},\\
  
  & publisher & = \{\{IEEE\}\},\\
  
  & doi       & = \{10.1109/SIBGRAPI54419.2021.00031\},\\
  \end{tabular}
  
\}
 }}}

\twocolumn
\linespread{1}\selectfont{}
\clearpage

%
\title{Improving Transferability of Domain Adaptation Networks Through Domain Alignment Layers}

\newif\iffinal
\finaltrue
\newcommand{\cmtid}{95}


\iffinal

\author{\IEEEauthorblockN{
Lucas F. A. Silva$^1$, 
Daniel C. G. Pedronette$^2$,
Fabio A. Faria$^1$, 
Jo\~{a}o P. Papa$^3$, and
Jurandy Almeida$^1$}\\
\IEEEauthorblockA{
    $^1$\textit{Instituto de Ci\^encia e Tecnologia, Universidade Federal de S\~{a}o Paulo - UNIFESP, S\~{a}o Jos\'{e} dos Campos, SP -- Brazil}\\
    Email: {\small\texttt{\{e.lucas, ffaria, jurandy.almeida\}@unifesp.br}}\\[1ex]
    $^2$\textit{Dept. of Statistics, Applied Mathematics and Computing, S\~{a}o Paulo State University - UNESP, Rio Claro, SP -- Brazil}\\
    Email: {\small\texttt{daniel.pedronette@unesp.br}}\\[1ex]
    $^3$\textit{Dept. of Computing, S\~{a}o Paulo State University - UNESP, Bauru, SP -- Brazil}\\
    Email: {\small\texttt{joao.papa@unesp.br}}}
}


%

\else
  \author{Sibgrapi paper ID: \cmtid \\ }
\fi

\maketitle

\begin{abstract}
Deep learning~(DL) has been the primary approach used in various computer vision tasks due to its relevant results achieved on many tasks. 
However, on real-world scenarios with partially or no labeled data, DL methods are also prone to the well-known domain shift problem. 
Multi-source unsupervised domain adaptation~(MSDA) aims at learning a predictor for an unlabeled domain by assigning weak knowledge from a bag of source models. 
However, most works conduct domain adaptation leveraging only the extracted features and reducing their domain shift from the perspective of loss function designs.
In this paper, we argue that it is not sufficient to handle domain shift only based on domain-level features, but it is also essential to align such information on the feature space. 
Unlike previous works, we focus on the network design and propose to embed Multi-Source version of DomaIn Alignment Layers~(MS-DIAL) at different levels of the predictor. 
These layers are designed to match the feature distributions between different domains and can be easily applied to various MSDA methods.
To show the robustness of our approach, we conducted an extensive experimental evaluation considering two challenging scenarios: digit recognition and object classification.
The experimental results indicated that our approach can improve state-of-the-art MSDA methods, yielding relative gains of up to +30.64\% on their classification accuracies.
\end{abstract}


\IEEEpeerreviewmaketitle

\section{Introduction}
\label{sec:introduction}

Machine learning algorithms, mainly the ones based on deep learning~(DL), have been the first approach to tackle a wide variety of problems, for they have achieved outstanding results and demonstrated a high generalization capacity on well-known benchmarks~\cite{IJCV_2021_Zhao, CVPR_2018_Xu, ICIAP_2017_Carlucci, ECCV_2020_Bucci, SIBGRAPI_2019_Santos, SIBGRAPI_2020_Santos}.
However, when applied to non-controlled environments, usually faced with real-world problems, DL methods can have their performance degraded to some extent. 
The first issue is that many real-world applications have no or few labeled data available in real-time~\cite{IJCV_2021_Zhao, TPAMI_2020_Carlucci, AAAI_2020_Zhao}. 
The most logical but naive approach concerns using similar sets of annotated data as a source to train the predictor for further inferring the unlabeled target data. 
However, even similar data can be drawn from different underlying distributions, \ie, they are obtained under different circumstances, exhibiting a phenomenon called \emph{domain shift}~\cite{TPAMI_2020_Carlucci, IJCV_2021_Zhao, ICCV_2019_Peng}.

The aforementioned shortcomings have been firstly tackled by the \emph{unsupervised domain adaptation}~(UDA) paradigm to adapt a single source domain to a single target domain on shallow machine learning methods, later extended to DL methods~\cite{ICCV_2019_Peng, CVPR_2018_Xu}.
UDA approaches, in general, can be divided into two main groups: (\textit{i}) methods that try to learn domain invariant features and (\textit{ii}) methods that minimize domain discrepancy, in a way that regularization terms are added to the loss function~\cite{ICIAP_2017_Carlucci}.
However, often it is possible to find more than one correlated dataset to the target task (\eg, digit recognition)~\cite{ICCV_2019_Peng, CVPR_2018_Xu}.
Additionally, it is necessary to consider the fact that data are not usually extracted from a single source of information, and can present as many as possible domains even within a same dataset, the so-called \emph{latent domains}~\cite{TPAMI_2021_Mancini}.
A common strategy to deal with more than one domain is to group them into a single source domain~\cite{ICML_2020_Wen, CVPR_2018_Xu}, which does not always hold effective results, once distinct domains may contribute in different ways to the transfer of knowledge.

The \emph{multi-source unsupervised domain adaptation}~(MSDA) has been proposed to mitigate the weakness of UDA methods by adapting a finite number of source domains to a single target domain. 
Indeed, this goal is more challenging, since we must deal with a shift between the source distributions. 
Each source domain can contribute with complementary information to fashion the knowledge about the target domain. 
Typically, they do not share the same label space~\cite{CVPR_2018_Xu}.
To reduce domain shift, most MSDA methods try to learn domain-invariant representations by introducing appropriate loss terms to penalize the discrepancy of features across domains~\cite{ICML_2020_Wen, CVPR_2018_Xu, ICCV_2019_Peng, AAAI_2020_Zhao}.

Although such methods are quite effective, we argue that it is not enough to deal with domain shift only based on the loss function and feature alignment at different levels of the network is also an indispensable part of domain adaptation.
Different from previous works based on designing loss functions solely, we focus on improving the transferability of the network by redesigning its architectural components.
For this, we propose to embed Multi-Source version of DomaIn Alignment Layers~(MS-DIAL)~\cite{WVC_2020_Silva} at different levels of any given DL model.
These layers are designed to match the feature distributions between different domains and can be easily applied to various MSDA methods.

Experiments were conducted on two challenging tasks using six popular benchmarks widely-used to evaluate MSDA methods: (\textit{i}) digit recognition using MNIST~\cite{IEEE_1988_Lecun}, MNIST-M~\cite{ICML_2015_Ganin}, SVHN~\cite{NIPS_2011_Netzer}, and \emph{Synthetic Digits}~\cite{JMLR_2016_Ganin} datasets; and (\textit{ii}) object classification on Office-31~\cite{ECCV_2010_Saenko} and Office-Home~\cite{CVPR_2017_Venkateswara} datasets.
We evaluated five different state-of-the-art MSDA methods and compared their results with and without using MS-DIAL.
The obtained results indicated that our approach yields significant effectiveness gains, reaching up to +30.64\% of relative gains on classification accuracies of the state-of-the-art MSDA methods. 
In addition, a visual analysis was conducted in order to highlight the benefits of our approach.

The remainder of this paper is organized as follows.
Section~\ref{sec:related-work} discusses related work.
Section~\ref{sec:dial} describes our approach and shows how it can be used for improving the transferability of any MSDA model. 
Section~\ref{sec:results} presents the experimental setup and reports our results.
Finally, we offer our conclusions and directions for future work in Section~\ref{sec:conclusions}.

\section{Related Work}
\label{sec:related-work}

UDA paradigm aims to transfer knowledge from a labeled source domain to an unlabeled target domain in order to learn a model well-performing on the target distribution.
Early UDA works have focused on shallow machine learning~(ML) methods but later they have shifted to DL methods~\cite{CVPR_2018_Xu}. 
On shallow methods, the main approaches rely on minimizing the discrepancy between the target domain and the source domain in order to obtain domain-invariant features, like the Transfer Component Analysis~(TCA)~\cite{TNN_2011_Pan} and Distribution-Matching Embeddings~(DME)~\cite{JMLR_2016_Baktashmotlagh}.

A common approach for DL methods consists in exploiting adversarial training to either transform source samples towards target samples or to overpower the feature extractor to fool the classifier by outputting features that are closer to the target domain, like Domain-Adversarial Neural Networks (DANN)~\cite{JMLR_2016_Ganin} and Weighted Maximum Mean Discrepancy (WMMD)~\cite{CVPR_2017_Yan}. 
Other methods try to align feature distributions by embedding domain-specific normalization layers into a neural network, such as DIAL~\cite{ICIAP_2017_Carlucci} and AutoDIAL~\cite{ICCV_2017_Carlucci}.

The aforementioned UDA methods try to reduce the domain shift from a single-source to a single-target. 
MSDA methods, on the other hand, face a challenging scenario of adopting more than one source domain to predict a single target domain. 
In the last years, many different MSDA approaches have been proposed in the literature, such as Moment Matching for Multi-Source Domain Adaptation (M3SDA)~\cite{ICCV_2019_Peng}, Deep CockTail Network (DCTN)~\cite{CVPR_2018_Xu}, Multiple Domain Matching Network (MDMN)~\cite{NIPS_2018_Li}, Multi-Source Domain Adversarial Networks (MDAN)~\cite{NIPS_2018_Zhao} and Domain Aggregation Networks (DARN)~\cite{ICML_2020_Wen}, Adversarial Domain Aggregation Networks (MADAN)~\cite{IJCV_2021_Zhao}, MultiDIAL~\cite{TPAMI_2020_Carlucci}, and MS-DIAL~\cite{WVC_2020_Silva}.

Except for these last two approaches, all the other MSDA works employ a multi-stream network, usually with one stream for each domain, some with different parameters for each stream and, in this case, each domain has an independent feature extractor and classifier; and others with parameters shared among the streams, usually by feature extractors from all domains, but each having its own classifier.
Usually, one stream is used to represent the task model and the others are used to align the target and source domains. 
In this way, a traditional task loss based on the labeled data and another alignment loss to tackle the domain shift problem are jointly optimized during the training phase. 

In this paper, we demonstrate that most of those MSDA methods can be improved by not just optimizing additional loss terms but also aligning feature distributions at different levels of the network.
For this, we use MS-DIAL~\cite{WVC_2020_Silva}, an extended version of DIAL~\cite{ICIAP_2017_Carlucci} for MSDA. 

\section{DIAL: DomaIn Alignment Layers}
\label{sec:dial}

Let ${\mathcal S} = \{\mathcal{S}_1, \mathcal{S}_2,\ldots,\mathcal{S}_M\}$ be a finite set of labeled source domains sharing the same set $\mathcal{Y}$ of categories with an unlabeled target domain $\mathcal{T}$. 
Each source domain $\mathcal{S}_i = \{ (\mathbf{x}_i^j, \mathbf{y}_i^j) \}_{j=1}^{N_i}$ refers to a set of tuples composed of $N_i$ samples $\mathbf{x}_i^j$ and their respective labels $\mathbf{y}_i^j$. 
Since we do not know the labels of the target domain beforehand, the set $\mathcal{T}=\{\mathbf{x}_T^j \}_{j=1}^{N_T}$ comprises the target samples only.

The source and target samples are drawn from distinct distributions. 
The distribution of the source domains can be estimated from the set of source samples and their labels. 
On the other hand, such information is unknown concerning the target domain. 
The final goal is to learn a function $f(\mathbf{x}_T;\theta)$ defined by a set of parameters $\theta$, whose input are the target samples $\mathbf{x}_T\in\mathcal{T}$. 
Such a function is expected to approximate the unknown target distribution and better classify samples in $\mathcal{T}$. 
This definition also holds to the single-source to single-target UDA problem by limiting the number of source domains, i.e., $M = 1$.

To solve this problem for a single source domain (\ie, $M=1$), Carlucci~\etal~\cite{ICIAP_2017_Carlucci} proposed to align feature distributions at different levels of a neural network through DomaIn Alignment Layers~(DIAL).
DIAL require one model only, shared across all domains, where each layer tries to bring all domain distributions to a canonical superposed one and, subsequently, to make few minor adjustments by performing a jointly linear transformation on all distributions. 
In this work, we adopt an extended version of DIAL that generalizes it for multiple source domains (\ie, $M>1$), called MS-DIAL~\cite{WVC_2020_Silva}.

MS-DIAL are based on Batch Normalization~(BN)~\cite{ICML_2015_Ioffe} layers, which normalizes batches based on the estimation of their first and second order statistics (\ie, mean and variance).
BN layers also use additional parameters to linearly transform the normalized features and let the model restore the batch representation.
MS-DIAL aligns all domain distributions to a canonical distribution, forwarding samples through a respective BN layer that is individual to each domain without performing the affine transformation. 
Finally, after the normalization process, MS-DIAL performs an affine transformation.

Note that MS-DIAL act in different ways on the training step and in the inference mode. 
We need to ensure that batches will comprise samples from all domains organized in the same order for training purposes. 
Concerning the inference mode, as we are interested in classifying samples from the target domain only, they are forwarded to their respective BN layer, driving MS-DIAL to use BN simply.

During training, source (and labeled) domain samples jointly with target (unlabeled) domain samples are fed to the neural network.
The loss function of our model is a weighted sum of a classification term (Equation~\ref{eq:supervisedloss}) and a distribution alignment term (Equation~\ref{eq:unsupervisedloss}). 
The classification term refers to the well-known cross-entropy loss function between the source predictions and their labels, as follows:

\begin{equation}
    \mathcal{L}_{\mathcal{S}}(\theta) = -\sum_{i=1}^{M} \sum_{k=1}^{N_i} \mathbf{y}_i^k \log f_i (\mathbf{x}_i^k; \theta),
    \label{eq:supervisedloss}
\end{equation}

where $f_i(\mathbf{x}_i^k;\theta)$ denotes the prediction function concerning source domain $\mathcal{S}_i$. 

The distribution alignment term minimizes the entropy of target samples in order to force the model to decide more confidently and is given by:

\begin{equation}
    \mathcal{L}_{\mathcal{T}}(\theta) = - \sum_{k=1}^{N_T} f_T( \mathbf{x}_T^k; \theta) \log f_T( \mathbf{x}_T^k;\theta),
    \label{eq:unsupervisedloss}
\end{equation}

in which $f_T(\mathbf{x}_T^k; \theta)$ denotes the prediction function for the target domain. The final loss function is computed as follows:

\begin{equation}
\mathcal{L}(\theta) = \mathcal{L}_{\mathcal{S}}(\theta) + \lambda \mathcal{L}_{\mathcal{T}}(\theta),
\end{equation}
where $\lambda$ weights the distribution alignment term.

We follow different strategies to include MS-DIAL on off-the-shelf methods: if the model has BN layers, we replace them by MS-DIAL and transfer their parameters to the new BN layers inside MS-DIAL; otherwise, if the model has no BN layers, we embed MS-DIAL after each convolutional and fully-connected layer to perform the feature alignment on each new representation, except on its final classification layer. This process is better described in Algorithm~\ref{alg:msdial-insertion}.

\begin{algorithm}[!htb]
\SetAlgoLined

\KwIn{DL model without MS-DIAL}
\KwOut{DL model with MS-DIAL}
\BlankLine
\tcc{The loop below iterates through all layers to verify if they contain BN layers.}
\BlankLine
\eIf{backbone model has BN layers}{
    \ForEach{layer $l$ of the backbone model}
    {
        \If{$l$ is BN Layer}
        {
            Replace it by MS-DIAL\;
        }
    }
}{
    \ForEach{layer $l$ of the backbone model}
    {
        \If{$l$ is a convolutional layer}
        {
            Replace it by a building block formed by the same convolutional layer but now followed by MS-DIAL\;
        }

        \ElseIf{$l$ is a fully-connected layer}
        {
            Replace it by a building block formed by the same fully-connected layer but now followed by MS-DIAL\;
        }
    }
}
\tcc{The affine parameters of the original BN layers, if present, are copied to MS-DIAL.}
\caption{Automatic MS-DIAL Insertion}
\label{alg:msdial-insertion}
\end{algorithm}

\begin{figure*}[!htb]
\centering
\subfloat[Digit recognition]{\includegraphics[width=0.565\textwidth]{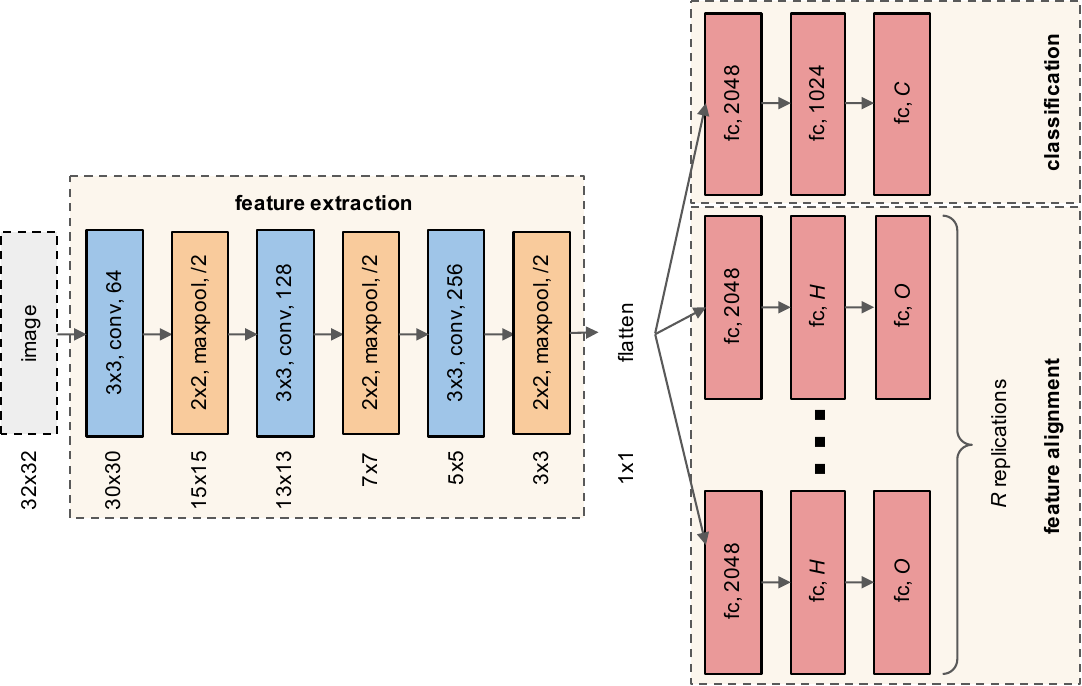}\label{fig:models-digits}}
\quad
\subfloat[Object Classification]{\includegraphics[width=0.4\textwidth]{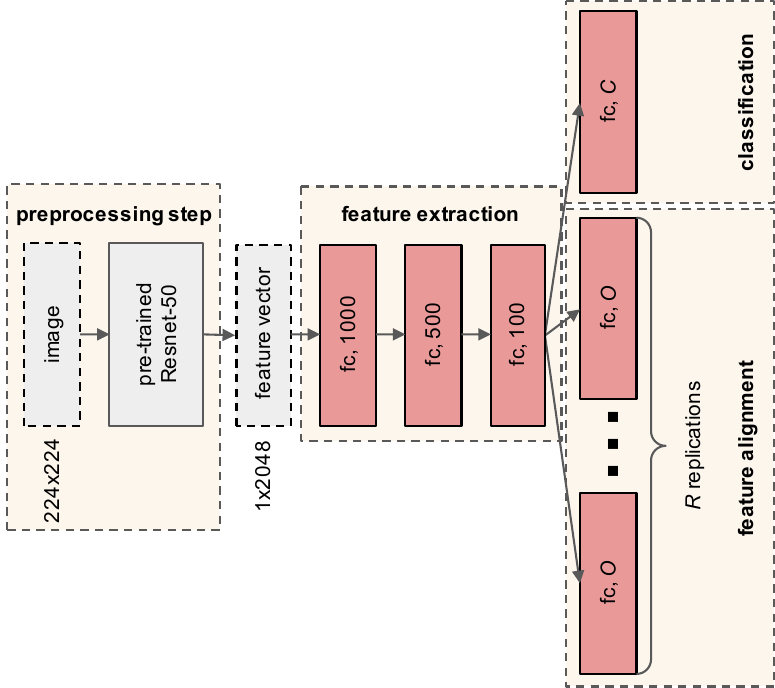}\label{fig:models-objects}}
\caption{Deep learning models used for each task, that are: (a) digit recognition, which, rv{except for the M3SDA}, is composed by a feature extractor with three convolutional layers and a classifier with three fully-connected~(FC) layers; and (b) object classification, where pre-computed features are extracted by a ResNet-50~\cite{CVPR_2016_He} model pre-trained on the ImageNet~\cite{IJCV_2015_Russakovsky} dataset and passed through four FC layers, in which the first three are for feature learning and the last is a classifier.
After feature extraction, the network is splitted into $R+1$ branches, where one performs classification and the others are for feature alignment.
The number $C$ of neurons in the output layer of the branch used for classification denotes the number of classes, which is 10 for digit recognition, 65 for object classification on the Office-Home dataset or 31 on the Office-31 dataset.
The number $R$ of branches used for feature alignment is the number of source domains ($M$) for DARN and MDAN; or $1$ for DANN, MDMN, and M3SDA.
The number $H$ of neurons in the hidden layer of such branches is $2048$ for DARN, MDAN, DANN, and MDMN; or $1024$ for M3SDA.
The number $O$ of neurons in the output layer of such branches is the number of domains ($M+1$) for MDMN; the number classes ($C$) for M3SDA; or $1$ for DARN, MDAN, and DANN.
Except for the output layers of all the branches, dropout is applied before each convolutional or FC layer, which are followed by ReLU activation.
}
\label{fig:models}
\end{figure*}

\section{Experiments and Results}
\label{sec:results}

In this section, we provide details about the experimental setup adopted in order to evaluate the proposed method as well as we report the obtained results.
A rigorous and extensive experimental evaluation was conducted on both small and large-scale datasets and the performance of state-of-the-art approaches was compared with and without using our approach.

\subsection{Datasets}
\label{ss.datasets}

The digit recognition was carried out on four different datasets, each of them related to a distinct domain, namely MNIST~\cite{IEEE_1988_Lecun}, MNIST-M~\cite{ICML_2015_Ganin}, SVHN~\cite{NIPS_2011_Netzer}, and Synth~\cite{JMLR_2016_Ganin}.
MNIST contains 70,000 grayscale images of handwritten digits with a resolution of 28x28 pixels, of which 60,000 are for training and 10,000 for testing.
MNIST-M~\cite{ICML_2015_Ganin} is composed by 59,001 training and 9,001 testing images with a resolution of 32x32 pixels obtained by modifying MNIST images with colored patches randomly extracted from the BSDS500 dataset~\cite{TPAMI_2011_Arbelaez}.
The Street View House Number~(SVHN)~\cite{NIPS_2011_Netzer} dataset contains 73,257 training and 26,032 testing images, in RGB color and size 32x32, of house numbers collected from Google Street View.
Synthetic Digits~(Synth)~\cite{JMLR_2016_Ganin} is a collection of 479,400 training and 9,553 testing images with size 32x32 generated from Windows fonts by varying position, orientation and background.
All these datasets contain 10 categories related to digits from '0' to '9'.

The object classification was conducted on the Office-31~\cite{ECCV_2010_Saenko} and Office-Home~\cite{CVPR_2017_Venkateswara} datasets.
Office-31~\cite{ECCV_2010_Saenko} is a standard benchmark in the MSDA literature. 
This dataset is composed of 4,652 images collected from Amazon.com or taken from an office environment using a Webcam or a DSLR camera and with varying lighting and pose changes. 
Those images comprise 31 classes from three different domains: Amazon, DSRL, and Webcam.
Office-Home~\cite{CVPR_2017_Venkateswara} is a large-scale benchmark widely-used for testing MSDA methods. 
This dataset is composed of 15,500 images collected from several search engines and online image directories.
They are distributed among 65 object categories and divided into 4 distinct domains: Art, Clipart, Product, and Real World.

\subsection{Experimental Protocol}
\label{sec:exp_protocol}

An essential point concerning DL works is their reproducibility, since most methods are highly affected by non-deterministic factors~\cite{ECCV_2020_Bucci}. 
A good practice is to re-run the source code with the parameters presented on the paper and compare its results with those obtained by the authors. 

For reproducibility purposes, our approach was implemented in PyTorch (version 1.4.0) upon the DARN~\cite{ICML_2020_Wen} implementation\footnote{\url{https://github.com/junfengwen/DARN} (As of July, 2021)}, whose source code is publicly available along with re-implementations of four state-of-the-art MSDA methods: DANN~\cite{JMLR_2016_Ganin}, MDAN~\cite{NIPS_2018_Zhao}, M3SDA~\cite{ICCV_2019_Peng}, and MDMN~\cite{NIPS_2018_Li}. 
Since our intent is to evaluate whether or not aligning domains on feature space may benefit existing MSDA methods, we first ran the original code and performed all its experiments, as suggested by Bucci~\etal~\cite{ECCV_2020_Bucci}. 
Next, we modified the original code with as minimum changes as possible in order to embed MS-DIAL into all the aforesaid methods\footnote{Our code is available at \url{https://github.com/LucasFernando-aes/MS-DIAL}}. 
Last but not least, we evaluated the use of MS-DIAL for MSDA on two different tasks: (\textit{i}) digit recognition and (\textit{ii}) object classification.

For a fair comparison, we adopt the same experimental protocol used by DARN~\cite{ICML_2020_Wen}.
In each experiment, one domain was chosen as the target and the rest was used as source domains. 
This process was repeated several times, each time with a different domain as the target.
Twenty replications were performed for each experiment in order
to ensure statistically sound results.
The reported results refer to the mean and standard error of the classification accuracies measured at the end of each experiment for all the replications.

We evaluated five different state-of-the-art MSDA methods, namely: \textbf{DANN}~\cite{JMLR_2016_Ganin}, \textbf{M3SDA}~\cite{ICCV_2019_Peng}, \textbf{MDAN}~\cite{NIPS_2018_Zhao}, \textbf{MDMN}~\cite{NIPS_2018_Li}, and \textbf{DARN}~\cite{ICML_2020_Wen}.
The purpose of our experiments is to compare their results with and without using MS-DIAL.
For reference, we report the results obtained using only \textbf{MS-DIAL}~\cite{WVC_2020_Silva}. Also, two additional baselines were considered: \textbf{SRC}, where a model was trained on a large set formed by merging all the source domains; and \textbf{TAR}, where a model was trained only on the target domain but having access to its true labels.
They can be seen as lower (SRC) and upper (TAR) bounds for the results that can be achieved by MSDA methods.

For digit recognition, we randomly chose 20000 training samples as the training set and 9000 testing samples as the testing set for each domain.
The network architecture used in such experiments is shown in Figure~\ref{fig:models-digits}.
Roughly speaking, it is a Convolutional Neural Network~(CNN) with three convolutional layers as a feature extractor on top of which is stacked a classifier with three fully-connected~(FC) layers, whose the output layer has 10 neurons, each corresponding to a particular digit. 
In addition to the classifier, three streams are used for feature alignment by DARN and MDAN; or only one stream in the case of DANN, MDMN, and M3SDA. 
These streams have a different amount of neurons depending on the MSDA method used, with 2048 neurons in the hidden layer and 1 in the output layer in the case of DANN, MDAN, and DARN; 2048 neurons in the hidden layer and 4 in the output layer for MDMN; and 1024 neurons in the hidden layer and 10 in the output layer for M3SDA.

For object classification, pre-computed features were first extracted by a ResNet-50~\cite{CVPR_2016_He} model pre-trained on the ImageNet~\cite{IJCV_2015_Russakovsky} dataset.
For this, we follow the same preprocessing step performed by He~\etal~\cite{CVPR_2016_He} to feed ResNet-50 models. 
During training, each image was resized to 256 pixels on its shortest side, and then a 224x224 crop was randomly sampled from the resulting image or its horizontal flip, normalizing it to the mean and standard deviation of the ImageNet dataset. 
In the inference mode, we first rescaled the image such that its shorter side was of length 256, then cropped out the central 224x224 patch from the resulting image and normalized it to the mean and standard deviation of the ImageNet dataset. 
The preprocessed images were passed through the network and 2048-dimensional feature vectors were extracted from its average pooling layer.
These feature vectors were passed through four FC layers, as shown in Figure~\ref{fig:models-objects}. The first three FC layers are for feature learning and have 1000, 500, and 100 neurons, respectively. 
The last layer is a classifier and has 31 output logits in the case of the Office-31 dataset and 65 for the Office-Home dataset. 
For feature alignment, three streams were used by DARN and MDAN; or only one stream in the case of DANN, MDMN, and M3SDA.
Each stream corresponds to single FC layer whose size is 1 for DARN, MDAN, and DANN; 4 for MDMN; and the number of classes for M3SDA (\ie, 31 for Office-31 or 65 for Office-Home).

Except for the output layers, dropout is applied before each convolutional or FC layer, which are followed by ReLU activation.
All the models were trained from scratch for 50 epochs using the Adadelta~\cite{ARXIV_2012_Zeiler} optimizer with a learning rate of $1.0$.
We used a mini-batch size of 128 for digit recognition and 32 for object classification.
The hyper-parameters adopted for each MSDA method are the same as described in~\cite{ICML_2020_Wen}.

\subsection{Ablation Study}

The key advantage of MS-DIAL is to align feature distributions.
To improve the model's confidence on unsupervised target samples, an entropy penalty is included in the loss function. 
If the entropy penalty is significant, the model may learn to make wrong predictions in order to minimize errors.

A critical part of MS-DIAL is to set the $\lambda$ hyper-parameter in order to balance entropy penalties with  classification errors. 
The former leads to more confident predictions for target samples whereas the latter takes knowledge from source domains. 

\begin{figure}[!htb]
\centering
\includegraphics[width=0.5\textwidth]{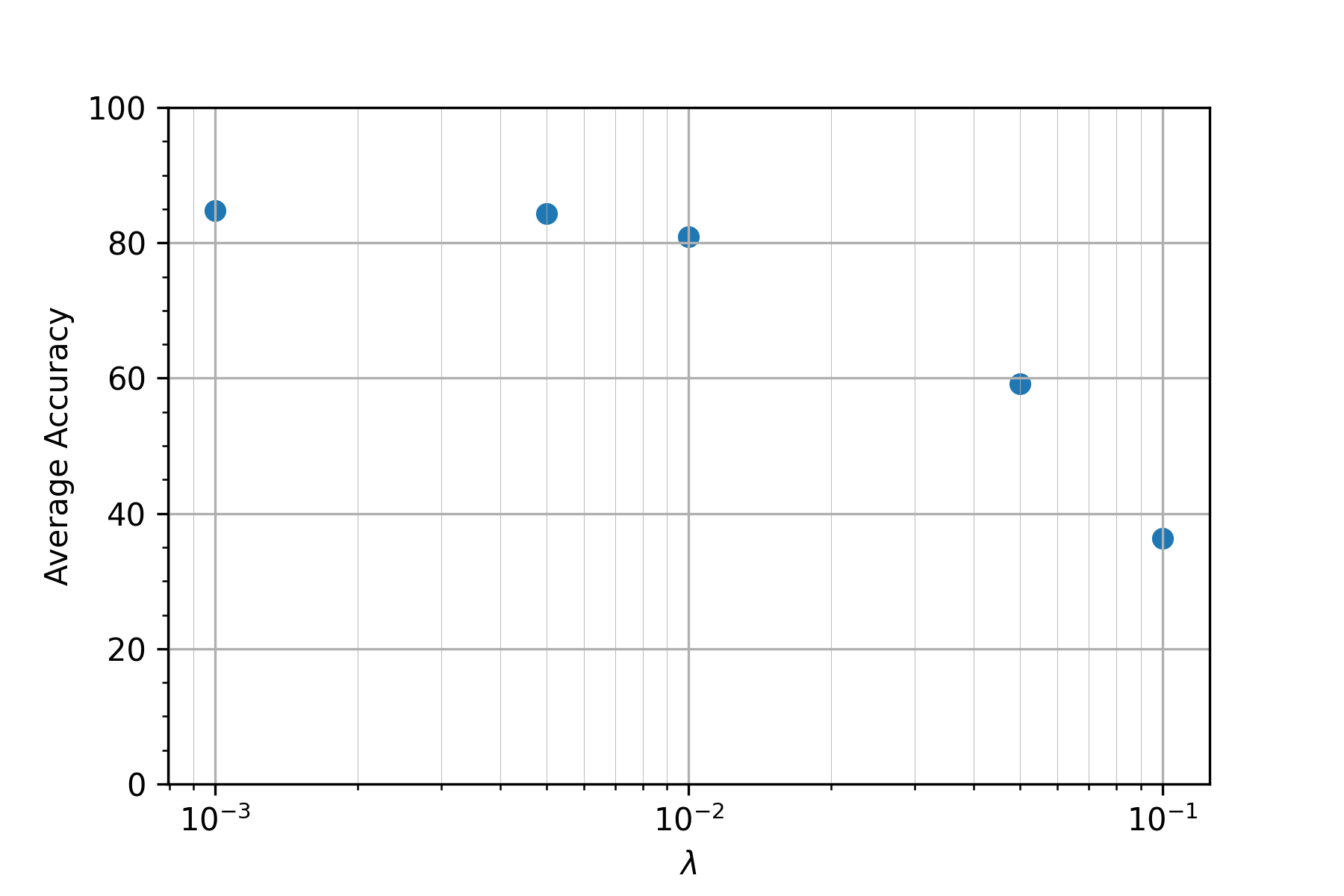}%
\vspace{-2mm}
\caption{Average accuracy results for different values of the $\lambda$ hyper-parameter.} 
\label{fig:entropy_weight}
\end{figure}

To assess the impact of the $\lambda$ hyper-parameter, we tested DARN with MS-DIAL on digit recognition task using five different values for the $\lambda$ hyper-parameter (i.e., $\lambda = \{0.001, 0.005, 0.01, 0.05, 0.1\}$) and compared its average accuracy for all the target domains, as shown in  Figure~\ref{fig:entropy_weight}. 
Notice that small values for the $\lambda$ hyper-parameter are better and, for this reason, we used $\lambda = 0.001$ in the next experiments.

\subsection{Quantitative Results}

Table~\ref{tab:digits_results} presents the results obtained for digit recognition.
When MNIST and Synth are used as target domain, probably the former by its simplicity (gray-scale images) and the latter because it is synthetically generated, we can observe a slight gain of performance between the original and MS-DIAL version, showing that the evaluated methods already can achieve good results for these adaptation tasks.
On the other hand, it does not hold for more difficult domains, presenting a bigger domain shift between them, like MNIST-M and SVHN, which have random patterns and colors and digit images coming from real world data, respectively.
When MS-DIAL are used, we can clearly see a gain of performance, even for DANN, which is a single-source to single-target UDA method.

Tables~\ref{tab:office31_results}~and~\ref{tab:officehome_results} present the results obtained for object classification on the Office-31 and Office-Home datasets, respectively.
For Office-31, the results obtained for the DSLR and Webcam domains were better than those for the Amazon domain.
However, in all cases, the use of MS-DIAL improved the performance of all the MSDA methods by a consistent margin, reaching, on average, +3\% of relative gains on the classification accuracy.
For Office-Home, as expected, the Clipart domain is the most difficult to be adapted from the other source domains, because it usually is a mix of digital artistic concepts with simplistic elements, making its results the worst among all the domains.
However, we can also see improvements of almost +13\% for DANN and 11\% for M3SDA with Clipart as target domain, and of almost +16\% for DANN and +15\% for M3SDA with Product as target domain, just by embedding MS-DIAL.
As we can observe, the use of MS-DIAL for aligning feature distributions can improve all the MSDA methods by a large margin, specially single-source to single-target UDA methods, like DANN, for which was obtained a relative gain of, on average, +30.64\% on its classification accuracy.

Finally, we can clearly notice that MS-DIAL is complementary to all the other MSDA methods, since for all the evaluated tasks its combination yielded superior results to those obtained by each method in isolation. For digit recognition, for instance, we can see improvements of, on average, +3\% in the results of M3SDA w/ MS-DIAL in relation to those of only MS-DIAL.

\begin{table*}[!htb]
\centering
\caption{Classification accuracy ($\%$) for digit recognition.}
\label{tab:digits_results}
\begin{tabular}{cccccccc}
\toprule
\multicolumn{2}{c}{\multirow{2}{*}{Methods}} & \multicolumn{4}{c}{Domains} & \multirow{2}{*}{Average} & \multirow{2}{*}{Relative Gain} \\ \cmidrule{3-6}
&                                     &       MNIST      & MNIST-M          & SVHN             & Synth            & \\ \midrule
\multicolumn{2}{c}{SRC}               & 96.48 $\pm$ 0.12 & 60.44 $\pm$ 0.41 & 68.32 $\pm$ 1.20 & 83.73 $\pm$ 0.31 & 77.24 $\pm$ 0.29 & - \\ \midrule
\multirow{2}{*}{DANN~\cite{JMLR_2016_Ganin}}   & w/o MS-DIAL & 96.70 $\pm$ 0.11 & 61.36 $\pm$ 0.37 & 67.42 $\pm$ 1.80 & 84.08 $\pm$ 0.32 & 77.39 $\pm$ 0.52 & \multirow{2}{*}{+5.71\%} \\
                        & w/ MS-DIAL  & 97.65 $\pm$ 0.08 & 64.85 $\pm$ 0.14 & 80.57 $\pm$ 0.37 & 84.15 $\pm$ 0.15 & 81.81 $\pm$ 0.11 \\ \midrule
\multirow{2}{*}{M3SDA~\cite{ICCV_2019_Peng}}  & w/o MS-DIAL & 96.44 $\pm$ 0.09 & 65.13 $\pm$ 0.29 & 76.96 $\pm$ 0.62 & 81.78 $\pm$ 0.26 & 80.08 $\pm$ 0.02 & \multirow{2}{*}{+7.07\%} \\
                        & w/ MS-DIAL  & 98.43 $\pm$ 0.03 & 72.58 $\pm$ 0.15 & 84.71 $\pm$ 0.29 & 88.98 $\pm$ 0.13 & 86.17 $\pm$ 0.08 \\ \midrule
\multirow{2}{*}{MDAN~\cite{NIPS_2018_Zhao}}   & w/o MS-DIAL & 97.27 $\pm$ 0.09 & 64.83 $\pm$ 0.30 & 76.51 $\pm$ 0.73 & 85.72 $\pm$ 0.14 & 81.03 $\pm$ 0.21 & \multirow{2}{*}{+3.04\%} \\
                        & w/ MS-DIAL  & 98.06 $\pm$ 0.07 & 69.03 $\pm$ 0.12 & 80.82 $\pm$ 0.40 & 86.04 $\pm$ 0.13 & 83.49 $\pm$ 0.10 \\ \midrule
\multirow{2}{*}{MDMN~\cite{NIPS_2018_Li}}   & w/o MS-DIAL & 97.21 $\pm$ 0.09 & 63.14 $\pm$ 0.33 & 76.74 $\pm$ 0.67 & 85.78 $\pm$ 0.17 & 80.72 $\pm$ 0.21 & \multirow{2}{*}{+4.14\%} \\
                        & w/ MS-DIAL  & 98.47 $\pm$ 0.04 & 68.48 $\pm$ 0.19 & 82.01 $\pm$ 0.70 & 87.29 $\pm$ 0.12 & 84.06 $\pm$ 0.20 \\ \midrule
\multirow{2}{*}{DARN~\cite{ICML_2020_Wen}}   & w/o MS-DIAL & 97.96 $\pm$ 0.03 & 67.97 $\pm$ 0.20 & 78.37 $\pm$ 0.34 & 86.61 $\pm$ 0.19 & 82.73 $\pm$ 0.12 & \multirow{2}{*}{+0.58\%} \\
                        & w/ MS-DIAL  & 97.94 $\pm$ 0.05 & 69.26 $\pm$ 0.33 & 79.21 $\pm$ 0.79 & 86.41 $\pm$ 0.12 & 83.21 $\pm$ 0.23 \\ \midrule
\multicolumn{2}{c}{MS-DIAL~\cite{WVC_2020_Silva}} & 98.45 $\pm$ 0.03 & 68.32 $\pm$ 0.19 & 81.82 $\pm$ 0.54 & 87.13 $\pm$ 0.08 & 83.93 $\pm$ 0.13 & -  \\ \midrule
\multicolumn{2}{c}{TAR}               & 99.04 $\pm$ 0.02 & 94.83 $\pm$ 0.08 & 87.48 $\pm$ 0.20  & 97.01 $\pm$ 0.06 & 94.59 $\pm$ 0.05 & -  \\
\bottomrule
\end{tabular}%
\end{table*}

\begin{table*}[!htb]
\centering
\caption{Classification accuracy ($\%$) for object classification on the Office-31 dataset.}
\label{tab:office31_results}
\begin{tabular}{ccccccc}
\toprule
\multicolumn{2}{c}{\multirow{2}{*}{Methods}} & \multicolumn{3}{c}{Domains} & \multirow{2}{*}{Average} & \multirow{2}{*}{Relative Gain} \\ \cmidrule{3-5}
                &                     &        Amazon    &         DSLR     & Webcam           &                      \\ \midrule
\multicolumn{2}{c}{SRC}               & 65.50 $\pm$ 0.23 & 93.92 $\pm$ 0.55 & 92.22 $\pm$ 0.28 & 83.88 $\pm$ 0.21 & - \\ \midrule
\multirow{2}{*}{DANN~\cite{JMLR_2016_Ganin}}   & w/o MS-DIAL & 66.02 $\pm$ 0.22 & 92.27 $\pm$ 0.62 & 93.06 $\pm$ 0.21 & 83.79 $\pm$ 0.21 & \multirow{2}{*}{+2.37\%} \\
                        & w/ MS-DIAL  & 67.00 $\pm$ 0.21 & 95.85 $\pm$ 0.45 & 94.48 $\pm$ 0.20 & 85.78 $\pm$ 0.19     \\ \midrule
\multirow{2}{*}{M3SDA~\cite{ICCV_2019_Peng}}  & w/o MS-DIAL & 65.34 $\pm$ 0.23 & 91.02 $\pm$ 0.69 & 91.06 $\pm$ 0.34 & 82.48 $\pm$ 0.28 & \multirow{2}{*}{+4.54\%} \\
                        & w/ MS-DIAL  & 67.01 $\pm$ 0.19 & 96.93 $\pm$ 0.43 & 94.74 $\pm$ 0.26 & 86.23 $\pm$ 0.18     \\ \midrule
\multirow{2}{*}{MDAN~\cite{NIPS_2018_Zhao}}   & w/o MS-DIAL & 66.53 $\pm$ 0.23 & 92.84 $\pm$ 0.55 & 92.17 $\pm$ 0.37 & 83.85 $\pm$ 0.25 & \multirow{2}{*}{+2.39\%} \\
                        & w/ MS-DIAL  & 67.37 $\pm$ 0.20 & 95.57 $\pm$ 0.49 & 94.62 $\pm$ 0.23 & 85.85 $\pm$ 0.22     \\ \midrule
\multirow{2}{*}{MDMN~\cite{NIPS_2018_Li}}   & w/o MS-DIAL & 63.03 $\pm$ 0.22 & 93.64 $\pm$ 0.52 & 92.84 $\pm$ 0.33 & 84.17 $\pm$ 0.24 & \multirow{2}{*}{+2.69\%} \\
                        & w/ MS-DIAL  & 67.94 $\pm$ 0.14 & 96.76 $\pm$ 0.50 & 94.60 $\pm$ 0.19 & 86.43 $\pm$ 0.18     \\ \midrule
\multirow{2}{*}{DARN~\cite{ICML_2020_Wen}}   & w/o MS-DIAL & 65.82 $\pm$ 0.33 & 93.69 $\pm$ 0.54 & 93.26 $\pm$ 0.30 & 84.26 $\pm$ 0.21 & \multirow{2}{*}{+2.59\%} \\
                        & w/ MS-DIAL  & 67.72 $\pm$ 0.18 & 96.93 $\pm$ 0.39 & 94.66 $\pm$ 0.26 & 86.44 $\pm$ 0.16     \\ \midrule
\multicolumn{2}{c}{MS-DIAL~\cite{WVC_2020_Silva}}       & 67.64 $\pm$ 0.18 & 96.08 $\pm$ 0.49 & 94.64 $\pm$ 0.25 & 86.12 $\pm$ 0.18 & - \\ \midrule
\multicolumn{2}{c}{TAR}               & 73.64 $\pm$ 0.32 & 95.34 $\pm$ 0.46 & 91.73 $\pm$ 0.46 & 86.90 $\pm$ 0.22 & - \\
\bottomrule
\end{tabular}%
\end{table*}

\begin{table*}[!htb]
\centering
\caption{Classification accuracy ($\%$) for object classification on the Office-Home dataset.}
\label{tab:officehome_results}
\begin{tabular}{cccccccc}
\toprule
\multicolumn{2}{c}{\multirow{2}{*}{Methods}} & \multicolumn{4}{c}{Domains} & \multirow{2}{*}{Average} & \multirow{2}{*}{Relative Gain} \\ \cmidrule{3-6}
                 &                 &           Art              & Clipart          & Product          & Real World       &                      \\ \midrule
\multicolumn{2}{c}{SRC}               & 42.53 $\pm$ 0.53 & 29.53 $\pm$ 0.21 & 55.85 $\pm$ 0.36 & 63.43 $\pm$ 0.23 & 47.84 $\pm$ 0.15 & - \\ \midrule
\multirow{2}{*}{DANN~\cite{JMLR_2016_Ganin}}   & w/o MS-DIAL & 42.68 $\pm$ 0.48 & 30.08 $\pm$ 0.38 & 55.68 $\pm$ 0.44 & 63.33 $\pm$ 0.33 & 47.94 $\pm$ 0.14 & \multirow{2}{*}{+30.64\%} \\
                        & w/ MS-DIAL  & 60.68 $\pm$ 0.47 & 43.83 $\pm$ 0.22 & 71.46 $\pm$ 0.21 & 74.54 $\pm$ 0.26 & 62.63 $\pm$ 0.14     \\ \midrule
\multirow{2}{*}{M3SDA~\cite{ICCV_2019_Peng}}  & w/o MS-DIAL & 48.22 $\pm$ 0.48 & 33.90 $\pm$ 0.17 & 57.14 $\pm$ 0.43 & 63.27 $\pm$ 0.28 & 50.63 $\pm$ 0.16 & \multirow{2}{*}{+25.81\%} \\
                        & w/ MS-DIAL  & 62.11 $\pm$ 0.79 & 44.42 $\pm$ 0.19 & 72.65 $\pm$ 0.16 & 75.60 $\pm$ 0.15 & 63.70 $\pm$ 0.20     \\ \midrule
\multirow{2}{*}{MDAN~\cite{NIPS_2018_Zhao}}   & w/o MS-DIAL & 55.92 $\pm$ 0.56 & 38.65 $\pm$ 0.22 & 66.48 $\pm$ 0.19 & 71.28 $\pm$ 0.21 & 58.08 $\pm$ 0.17 & \multirow{2}{*}{+7.49\%} \\
                        & w/ MS-DIAL  & 60.75 $\pm$ 0.44 & 44.28 $\pm$ 0.22 & 70.66 $\pm$ 0.18 & 74.03 $\pm$ 0.16 & 62.43 $\pm$ 0.12     \\ \midrule
\multirow{2}{*}{MDMN~\cite{NIPS_2018_Li}}   & w/o MS-DIAL & 57.49 $\pm$ 0.43 & 38.66 $\pm$ 0.25 & 68.34 $\pm$ 0.28 & 72.52 $\pm$ 0.22 & 59.25 $\pm$ 0.17 & \multirow{2}{*}{+7.20\%} \\
                        & w/ MS-DIAL  & 62.76 $\pm$ 0.55 & 44.46 $\pm$ 0.20 & 71.91 $\pm$ 0.18 & 74.91 $\pm$ 0.14 & 63.51 $\pm$ 0.16     \\ \midrule
\multirow{2}{*}{DARN~\cite{ICML_2020_Wen}}   & w/o MS-DIAL & 57.59 $\pm$ 0.55 & 40.21 $\pm$ 0.23 & 69.40 $\pm$ 0.22 & 73.56 $\pm$ 0.24 & 60.19 $\pm$ 0.16 & \multirow{2}{*}{+4.84\%} \\
                        & w/ MS-DIAL  & 61.17 $\pm$ 0.38 & 42.84 $\pm$ 0.14 & 73.02 $\pm$ 0.16 & 75.38 $\pm$ 0.16 & 63.10 $\pm$ 0.12     \\ \midrule
\multicolumn{2}{c}{MS-DIAL~\cite{WVC_2020_Silva}}& 62.25 $\pm$ 0.47 &   44.28 $\pm$ 0.15 & 72.27 $\pm$ 0.18 & 74.90 $\pm$ 0.21 & 63.42 $\pm$ 0.15 & - \\ \midrule
\multicolumn{2}{c}{TAR}            & 57.68 $\pm$ 0.54 & 42.39 $\pm$ 0.31 & 79.54 $\pm$ 0.26 & 73.58 $\pm$ 0.22 & 63.29 $\pm$ 0.23 & - \\
\bottomrule
\end{tabular}%
\end{table*}

\subsection{Visualization Analysis}
\label{sss.umap}

In order to enrich the discussion about the proposed approach, we employed dimensionality reduction methods to represent the impact of the proposed method on a 2-D projection of feature space. 
The analysis was performed on all the digits datasets, using the well-known UMAP~\cite{ARXIV_2018_McInnes} algorithm.
Due to space limitations, we chose to report the results only for MDMN, comparing the projections produced with and without using MS-DIAL.
However, similar results were observed for all the MSDA methods.

\begin{figure*}[!htb]
\centering
\includegraphics[width=0.9\textwidth]{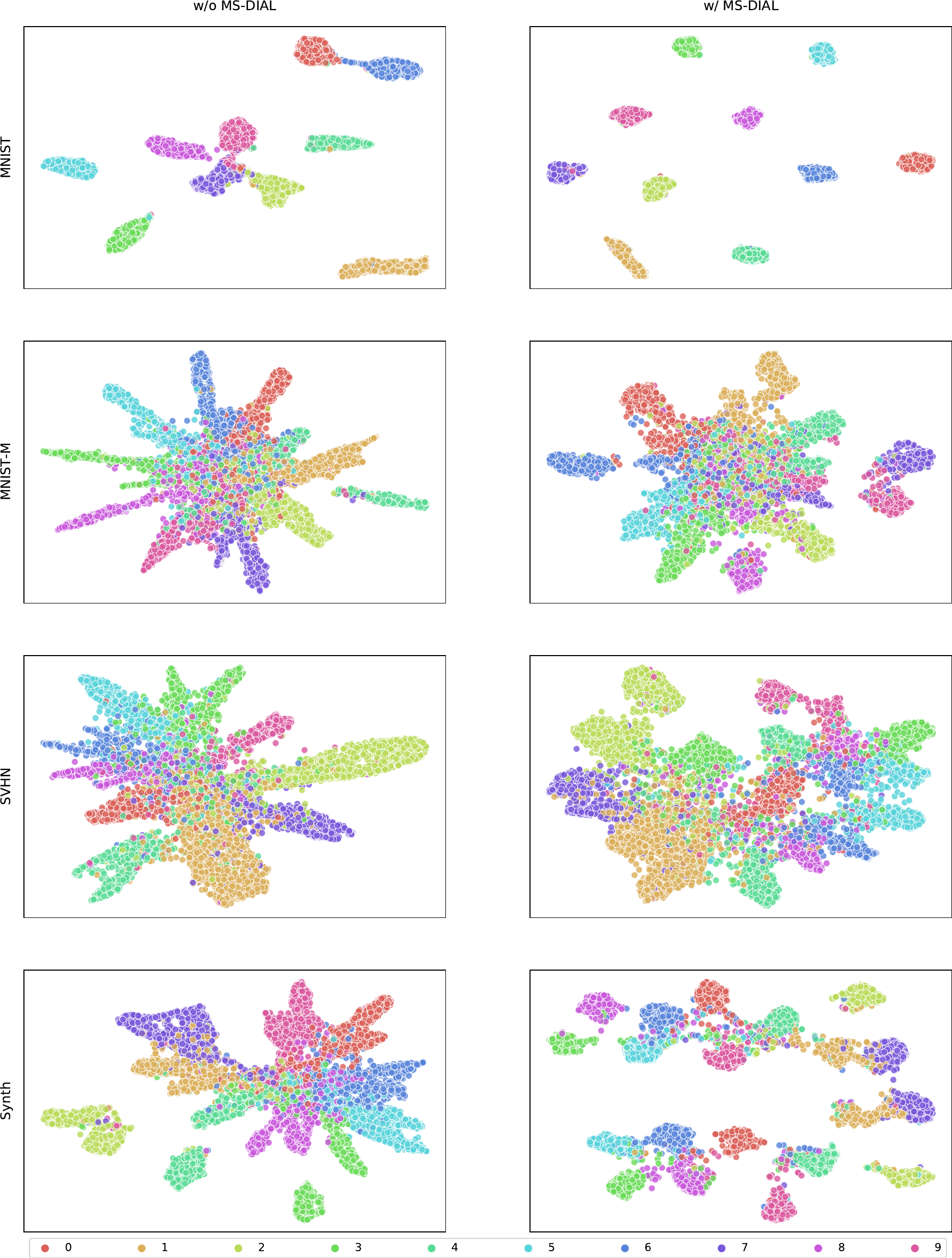}%
\caption{UMAP~\cite{ARXIV_2018_McInnes} dimensional reduction of input features in final classification layers (100-dimensional vectors) to 2-dimensional vectors.
Columns refer to, respectively, MDMN original method and MDMN adapted with MS-DIAL, on digit recognition task, with each row indicating the target domain.
}
\label{fig:umap}
\end{figure*}

Figure~\ref{fig:umap} presents the visualizations of the application of UMAP on the digits datasets.
In most situations, an evident increase in the separability among classes can be observed. 
We can highlight the results for MNIST, presented in the first row, where the class representations are clearly apart when using the proposed approach. 

\section{Conclusions}
\label{sec:conclusions}

In this paper, we focused on improving the transferability of DL models with simple and efficient network layers easily pluggable into the network backbones of existing MSDA methods.
More specifically, we proposed to embed MS-DIAL~\cite{WVC_2020_Silva} at different levels of any given DL model.
The main contribution of MS-DIAL is to perform the feature alignment on each new representation generated along the network.

Our approach was validated on digit recognition and object classification tasks.
Six popular benchmarks widely-used to evaluate MSDA methods and five different state-of-the-art MSDA methods were considered in our experiments.
Extensive experimental results demonstrated that MS-DIAL significantly boosts existing MSDA methods, leading to relative gains of up to +30.64\% on their classification accuracies.

As future work, we intend to evaluate the use of MS-DIAL with other MSDA methods.
Also, we want to augment MS-DIAL by exploring latent domain discovery strategies.
In addition, we also plan to extend our ideas to other types of domain adaptation problems, like domain generalization. 


\iffinal
\section*{Acknowledgments}
This research was supported by the São Paulo Research Foundation - FAPESP (grants
2013/07375-0, 
2014/12236-1 
2017/25908-6, 
2018/15597-6, 
2019/07665-4, and 
2020/08770-3) 
and the Brazilian National Council for Scientific and Technological Development - CNPq (grants
307066/2017-7, 
427968/2018-6, 
309439/2020-5, and 
314868/2020-8). 
\fi


%
%





\end{document}